\documentclass{article}

\usepackage{PRIMEarxiv}

\usepackage[utf8]{inputenc} 
\usepackage[T1]{fontenc}    
\usepackage{hyperref}       
\usepackage{url}            
\usepackage{booktabs}       
\usepackage{amsfonts}       
\usepackage{nicefrac}       
\usepackage{microtype}      
\usepackage{lipsum}
\usepackage{xcolor} 
\usepackage{fancyhdr}       
\usepackage{graphicx}       
\graphicspath{{media/}}     

\usepackage{amsmath,amssymb,amsfonts}
\usepackage{algorithmic}
\usepackage{graphicx}

\usepackage{textcomp}
\usepackage{bm}
\usepackage{graphicx}
\usepackage{tabularx}
\usepackage{scalerel}

\usepackage{multirow}
\usepackage{float}
\usepackage[ruled,vlined,linesnumbered]{algorithm2e}

\usepackage{amsmath}
\usepackage{url}

\newcommand{\B}{\bfseries} 

\title{Bio-inspired fine-tuning for selective transfer learning in image classification}

\author{
  Ana Davila \\
  Institutes of Innovation for Future Society \\
  Nagoya University \\
  Nagoya, Japan\\
  \texttt{adavila@nagoya-u.jp} 
  \And
  Jacinto Colan \\
  Dept. of Micro-Nano Mechanical Science and Engineering \\
  Nagoya University \\
  Nagoya, Japan\\
  \texttt{jcolan@nagoya-u.jp} 
  \And
  Yasuhisa Hasegawa \\
  Institutes of Innovation for Future Society \\
  Nagoya University \\
  Nagoya, Japan\\
  \texttt{hasegawa@nagoya-u.jp}
}

\usepackage[absolute,overlay]{textpos}
\begin{document}

\begin{textblock*}{18cm}(1.5cm,1cm) 
    \centering
    \footnotesize
    \textcolor{gray}{Published in \textit{IEEE Access}, vol. 13, pp. 129234--129249, 2025. DOI: 10.1109/ACCESS.2025.3587524}
\end{textblock*}

\maketitle

\begin{abstract}
Deep learning has significantly advanced image analysis across diverse domains but often depends on large, annotated datasets for success. Transfer learning addresses this challenge by utilizing pre-trained models to tackle new tasks with limited labeled data. However, discrepancies between source and target domains can hinder effective transfer learning. We introduce BioTune, a novel adaptive fine-tuning technique utilizing evolutionary optimization. BioTune enhances transfer learning by optimally choosing which layers to freeze and adjusting learning rates for unfrozen layers. Through extensive evaluation on nine image classification datasets, spanning natural and specialized domains such as medical imaging, BioTune demonstrates superior accuracy and efficiency over state-of-the-art fine-tuning methods, including AutoRGN and LoRA, highlighting its adaptability to various data characteristics and distribution changes. Additionally, BioTune consistently achieves top performance across four different CNN architectures, underscoring its flexibility. Ablation studies provide valuable insights into the impact of BioTune's key components on overall performance. The source code is available at \url{https://github.com/davilac/BioTune}.
\end{abstract}


\section{Introduction}
\label{sec:1}

Despite the success of deep learning in various image analysis applications \cite{kamilaris18deep,angermueller16deep,fozilov23endoscope,litjens17survey}, the requirement for large annotated datasets remains a significant challenge in many specialized fields that require expert annotation or involve sensitive information, such as medical imaging \cite{yamada24multimodal, yamada23task}. This data scarcity motivates the development of techniques that can leverage existing labeled data more effectively \cite{zhang24fromsample}.

Transfer learning offers a practical solution to the limited data challenge by adapting pre-trained models from data-rich domains to specific target tasks. This approach reduces the need for extensive labeled datasets while improving training efficiency and classification performance. However, effective transfer learning faces substantial challenges when source and target domains differ significantly, potentially leading to negative transfer or catastrophic forgetting, where the adapted model performs worse than baseline training.
Fine-tuning, a common transfer learning approach, involves adjusting pre-trained model parameters using task-specific data. Although this method exploits useful feature representations, it presents several technical challenges, including the potential deterioration of pre-trained features and decreased performance on out-of-distribution data, especially when domains vastly differ \cite{kumar22fine}. A common practice in fine-tuning is the selective adjustment of layers. However, selecting which layers to fine-tune adds complexity. Traditionally, efforts have focused on training only the terminal layers, but recent studies suggest that adaptive layer selection strategies might be more effective \cite{lee23surgical}. Furthermore, additional fine-tuning parameters such as learning rates and weight decay significantly impact model performance and stability \cite{alshalali18fine}, often requiring domain-specific expertise \cite{nguyen21fine}.

To address these challenges, the fine-tuning process can be framed as an optimization problem, thereby enabling the application of nature-inspired algorithms. These algorithms have demonstrated efficacy in tackling complex optimization tasks, including constrained inverse kinematics \cite{davila24realtime}, hyperparameter tuning \cite{vincent23improved}, and feature selection \cite{nssibi23advances}. Among these, Evolutionary Algorithms (EA) and Swarm Intelligence (SI) approaches each have their own distinct strengths. EAs outperform in maintaining population diversity, enhancing solution robustness, while SI algorithms are known for their efficiency and simplicity \cite{vesterstrom04comparative}. Hybrid algorithms that combine elements from both EAs and SI aim to leverage their collective strengths, balancing the exploration of new solutions with the exploitation of known good configurations \cite{grosan07hybrid}.

In this paper, we introduce an adaptive fine-tuning method using an improved evolutionary algorithm that combines genetic operators with population momentum to explore fine-tuning configurations. The proposed approach automatically determines layer freezing strategies and optimizes learning rates for active layers, achieving efficient exploration through systematic evaluation of target dataset subsets during each evolutionary generation.
We evaluate this methodology using pre-trained convolutional neural networks (CNNs) across standard and specialized image classification tasks. Results demonstrate improved accuracy and efficiency compared to existing approaches, while providing insights into task-specific adaptations and their effects on CNN representations.
The main contributions of this paper are as follows.

\begin{itemize}
    \item  A novel evolutionary-based adaptive fine-tuning approach that dynamically optimizes learning rates and layer freezing strategies across diverse datasets and architectures.
    \item Comprehensive comparison against existing fine-tuning methods across nine image classification datasets and four CNN architectures, demonstrating improved performance and adaptability.
    \item Ablation studies providing insights into the impact of key components of our proposed algorithm on overall performance.
\end{itemize}

To facilitate reproducibility and further research, the implementation of the proposed algorithm is available at \url{https://github.com/davilac/BioTune}
. The remainder of this paper is organized as follows. Section~\ref{sec:2} reviews the related work on transfer learning and adaptive fine-tuning. Section~\ref{sec:3} describes the design and implementation of the proposed hybrid evolutionary fine-tuning algorithm. The experimental setup is presented in Section~\ref{sec:4}. In Section~\ref{sec:5}, we discuss the results and comparisons with existing techniques. Finally, in Section~\ref{sec:6}, we conclude the paper by summarizing the implications and limitations of our approach and suggest some future directions.

\section{Related works}
\label{sec:2}

\subsection{Transfer learning}
\label{sec:2.1}

Transfer learning enables the application of pre-trained models to new domains and tasks by leveraging knowledge gained from source domains with abundant labeled data \cite{pan10survey}. This approach reduces the need for extensive labeled data in target domains, accelerates training, and minimizes overfitting \cite{hussain19study}. The technique has proven particularly effective in computer vision and natural language processing applications \cite{zhuang21comprehensive}.

Transfer learning approaches can be broadly categorized into domain generalization and inductive transfer learning. Domain generalization focuses on developing robust models that can generalize directly to target distributions without requiring labeled target data \cite{peters16causal, arjovsky19invariant}. In contrast, inductive transfer learning utilizes labeled data from the target domain to refine pre-trained models, consistently demonstrating superior performance compared to domain generalization and unsupervised adaptation methods \cite{rosenfeld22domain, kirichenko22last}. Within the inductive framework, researchers employ either multi-task learning or sequential learning strategies. Multi-task learning aims to identify common latent features beneficial across multiple tasks simultaneously, requiring a balanced data distribution from both source and target domains. Sequential learning follows a two-step approach of pre-training and adaptation, where pre-training generates a general representation model applicable to various tasks, usually requiring a large and diverse dataset. Adaptation refines this model for the specific target task by reusing the pre-trained parameters.

The adaptation step can be generally performed in two main ways: feature-extraction and fine-tuning \cite{zhuang21comprehensive}. Feature-extraction methods keep the pre-trained weights fixed and use them to extract relevant features for the target task. These features are then used as input for a classifier that is trained on the specific application \cite{shi19deep,nogueira17towards}. This approach enables the reuse of specialized models and reduces computational costs when the same features can be used for multiple tasks. Fine-tuning methods, on the other hand, update the pre-trained representations according to the target domain. They use the pre-trained weights as initial values for the target task model, which has the same architecture as the pre-training model alongside some task-specific layers. They allow the gradients to backpropagate to the pre-trained parameters, and thus modify them based on the target application. Fine-tuning usually requires minimal changes in model architecture and allows for the adaptation of a single general-purpose model to multiple tasks. Previous works have evaluated both methods on different tasks, showing that fine-tuning works better when the tasks are related (e.g., source and target tasks are both supervised classification tasks), while feature extraction works better when the tasks are different (e.g. from image classification to image segmentation) \cite{peters19tune, kornblith19dobetter}. 

However, the success of transfer learning relies on the assumption of alignment between the source and target domains \cite{wang18deep}. When the alignment assumption is violated due to distribution shifts, negative transfer can occur, where a pre-trained model hinders adaptation, leading to worse performance than training from scratch \cite{chen19catastrophic}. These distribution shifts occur when the data distributions in the source and target domains differ \cite{quinonero08dataset}, due to  inherent domain differences or natural variations within a single domain \cite{taori20measuring}. These distribution shifts can compromise the reliability of transferred models, highlighting the need for transfer learning approaches capable of effectively mitigating distribution shifts and adapting to new domains or tasks \cite{wang20transfer}.

\subsection{Adaptive fine-tuning strategies}
\label{sec:2.2}

\begin{figure}[t!] 
  \centering
  \includegraphics[width=0.5\linewidth]{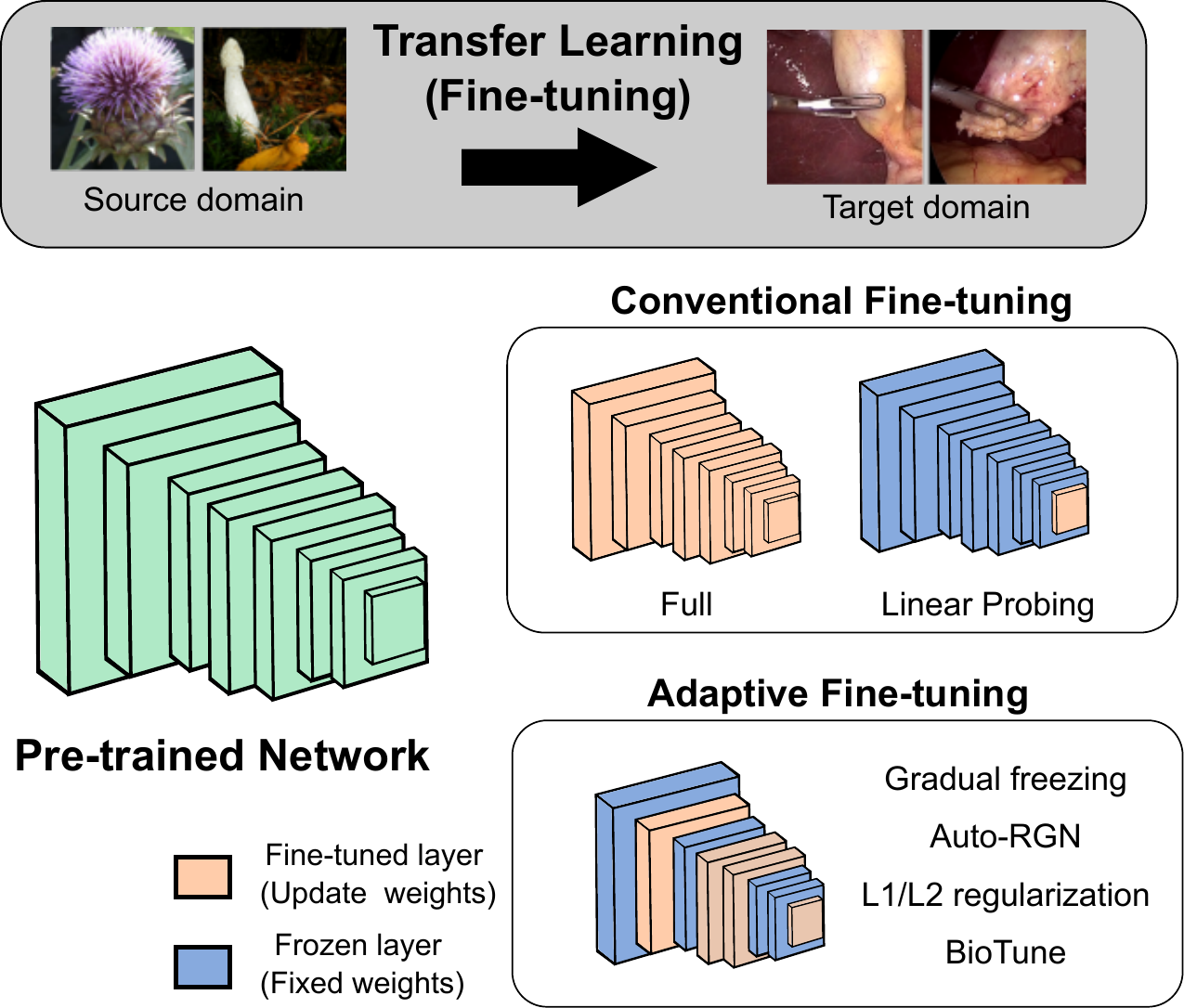}
  \caption{Overview of the transfer learning process via fine-tuning, illustrating the adaptation of a pre-trained model to a target domain.}
  \label{fig:1}
\end{figure}

Various fine-tuning strategies have been proposed to optimize transfer learning from source to target domains, as illustrated in Figure~\ref{fig:1}. One common approach is linear probing, which involves freezing the pre-trained network layers and only training a new classifier layer on top using the target domain data \cite{girshick14rich}. This method preserves the learned feature representations while adapting the final classification decision boundary to the new task.  However, linear probing can be suboptimal when the target task requires more substantial feature adaptation or when the source domain features are not sufficiently relevant for the target task. In contrast, full fine-tuning involves updating all the model's parameters, including both the pre-trained layers and the new classifier, to adapt to the target domain. However, this approach can lead to significant gradient flow through the entire network, potentially causing catastrophic forgetting of useful features learned during pre-training. Although full fine-tuning often achieves higher in-distribution accuracy on the target domain, studies have shown it can underperform compared to linear probing when evaluating on out-of-distribution samples with substantial distribution shifts \cite{kumar22fine}. This suggests that full fine-tuning may not always be the best approach, particularly when the target domain differs significantly from the source domain.

To address the challenges of distribution shifts, overfitting, and computational costs in transfer learning, several strategies have been proposed to enhance the fine-tuning process. A primary approach involves selecting relevant samples from the source or target distribution to minimize the domain gap between source and target tasks, thereby preventing negative transfer. For instance, Zhu et al. \cite{zhu20learning} introduced a meta-learning optimization method that identifies and selects informative samples from the target distribution specifically for fine-tuning. Similarly, Cui et al. \cite{cui18large} developed a domain similarity metric to identify and select source samples most similar to the target distribution, subsequently pre-training a CNN on these selected samples before fine-tuning on the target domain. Ge et al. \cite{ge17borrowing} proposed selective joint fine-tuning, where they identify a subset of training images from the source domain whose low-level characteristics align with the target training set, and jointly fine-tune shared convolutional layers for both tasks. Although sample selection techniques can improve data efficiency and performance, relying solely on these methods may not fully address distribution shifts caused by differing tasks or domains. Additionally, these approaches can introduce additional computational costs or require extra annotations.

As pre-trained models continue to grow in size, fully fine-tuning them on downstream tasks with limited data carries a high risk of overfitting. To mitigate this, some studies have explored regularization techniques and learning rate adjustments. Li et al. \cite{li18explicit} investigated several regularization schemes that encourage similarity between the fine-tuned model and the initial pre-trained model. They found that an explicit inductive bias towards the initial model, such as an L2 penalty, serves as an effective baseline for transfer learning tasks. In another study, Li et al. \cite{li20rethinking} demonstrated that maintaining a lower learning rate during fine-tuning compared to pre-training reduces the risk of catastrophic forgetting.

Selective fine-tuning is another approach that involves determining which layers of a pre-trained model should be fine-tuned to optimize knowledge extraction while keeping other layers frozen. This approach is based on the premise that different layers capture varying levels of feature abstraction and complexity. Depending on the similarity between source and target tasks, some layers may be more suitable for fine-tuning than others.  It is commonly assumed that initial layers capture more generic features applicable to a wide range of tasks, while later layers learn task-specific features \cite{yosinski14howtransferable, azizpour16factors}. Consequently, some methods propose freezing the initial layers and only fine-tuning the final ones to preserve general features while adapting to target-specific characteristics \cite{long15learning}. Additionally, strategies like gradually unfreezing layers from top to bottom during fine-tuning have been proposed as a form of regularization, encouraging new parameters to stay close to their pre-trained values \cite{howard18universal, romero20targeted}. However, this approach may not be optimal for all distribution shifts, as certain layers might be more relevant or sensitive to specific types of changes \cite{guo19spottune}. 

Adaptive fine-tuning methods aim to refine model parameters based on training performance rather than adhering to predefined strategies. Tajbakhsh et al. \cite{tajbakhsh16convolutional} proposed a layer-wise relevance propagation method to identify the most relevant layers for fine-tuning in medical image analysis. Chougrad et al. \cite{chougrad18deep} developed a layer selection criterion based on the correlation between the activations of each layer and the target labels in the detection of diabetic retinopathy. Meta-heuristic approaches have also been utilized to optimize layer selection for fine-tuning. Vrbancic et al. \cite{vrbancic20transfer} used differential evolution to identify the optimal combination of layers to fine-tune, demonstrating improved performance in osteosarcoma detection from medical images compared to full fine-tuning and last-layer fine-tuning. Similarly, Nagae et al. \cite{nagae22automatic} used a genetic algorithm to automatically select the most effective layers for transfer learning, incorporating a quantitative evaluation method based on optimal transport distance to assess layer effectiveness. Shen et al. \cite{shen21partial} propose a method to transfer partial knowledge by selectively freezing or fine-tuning particular layers in the base model. They utilize an evolutionary search-based method to efficiently determine which layers should be fine-tuned and to assign appropriate learning rates. However, these approaches often constrain the exploration space with predefined hyperparameters, such as discrete learning rates, and do not account for the interdependencies between layer selection and hyperparameter adjustments. Additionally, they may not scale effectively to larger datasets due to the computational expense of each iteration. Lee et al. \cite{lee23surgical} proposed AutoRGN, which dynamically modifies the learning rate of each layer based on the relative ratio of gradient norm to parameter norm. Evaluated on image classification tasks with various distribution shifts, AutoRGN outperformed full fine-tuning, gradual unfreezing, and regularization-based fine-tuning approaches. This technique has also been applied to other nonlinear optimization tasks using gradient-based methods \cite{davila23gradient, colan24variable}. Ro et al. \cite{ro21autolr} introduced an algorithm that dynamically adjusts learning rates and prunes weights for each layer based on their role and importance.

Parameter-efficient fine-tuning (PEFT) methods aim to reduce the number of trainable parameters by incorporating learnable adaptation modules into the frozen backbone of the model. LoRA \cite{hu21lora} integrates low-rank matrices into the weight updates of each layer, significantly decreasing the number of trainable parameters while often maintaining comparable performance. This approach is particularly beneficial for large models in transformer architectures, enabling efficient adaptation without the computational demands of full fine-tuning. Similarly, Chen et al. \cite{chen24convadapter} proposed ConvAdapter, a bottleneck structure for fine-tuning ConvNets that utilizes two concatenated convolutional layers with a non-linearity in between. The first layer downsamples the channel dimension, and the second projects it back, maintaining a similar receptive field to the adapted backbone.

In contrast to prior methods, our approach introduces a novel evolutionary strategy that combines selective and adaptive fine-tuning to optimize transfer learning. By simultaneously freezing layers and dynamically adjusting learning rates, our method aims to enable a more flexible and efficient adaptation to diverse tasks and domains.

\section{Proposed method}
\label{sec:3}

\begin{figure}[t!]
  \centering
  \includegraphics[width=0.95\textwidth]{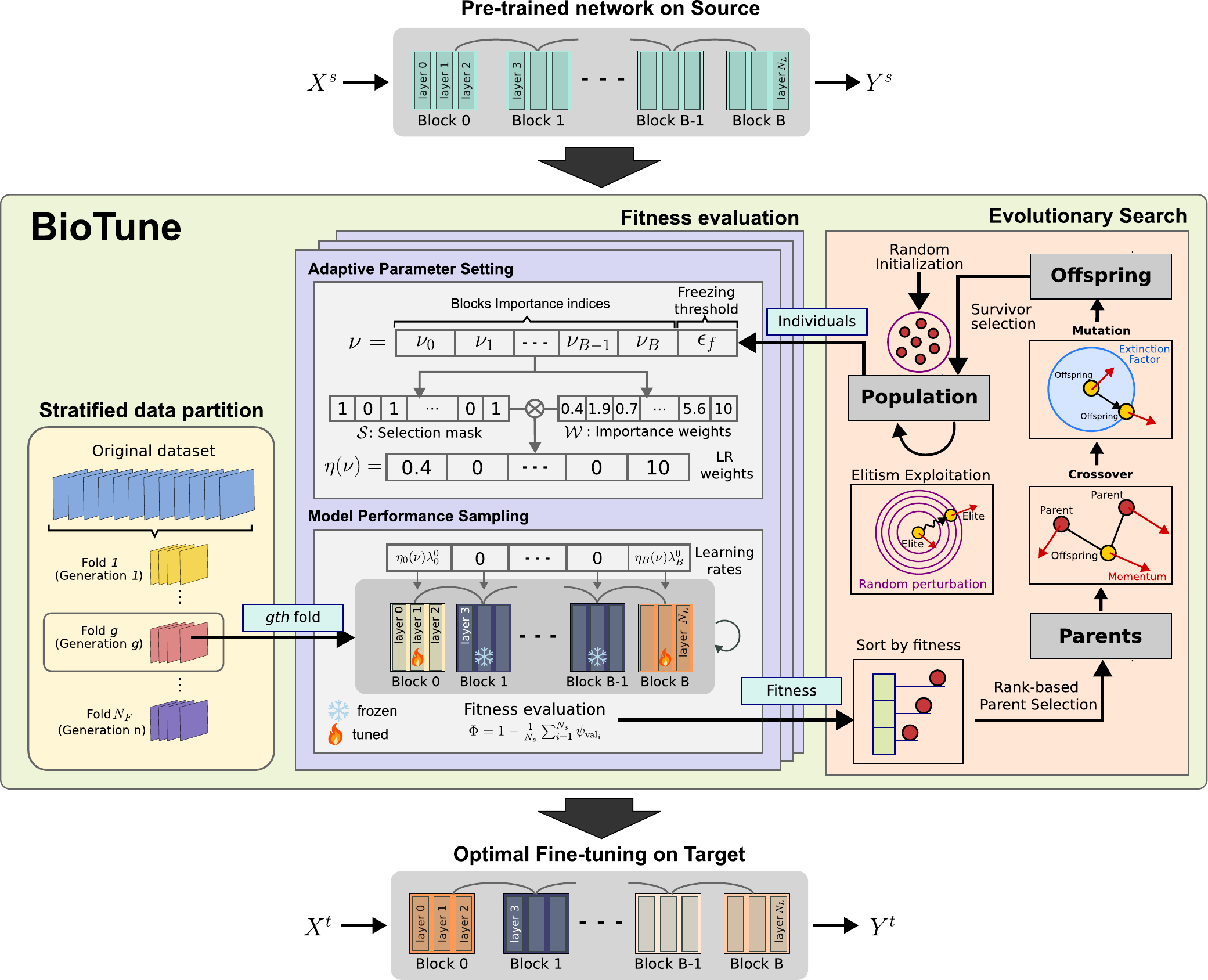}
  \caption{Overview of BioTune.}
  \label{fig:2}
\end{figure}

We propose a novel fine-tuning framework for selective and adaptive fine-tuning, which simultaneously optimizes the set of layers to fine-tune and their corresponding learning rates to maximize classification performance. Our approach leverages an improved evolutionary optimization strategy, incorporating population momentum to guide the exploration process \cite{starke19memetic}. An overview of the proposed method is presented in Figure \ref{fig:2}, with a detailed description provided in Algorithm \ref{alg:1}. The BioTune method comprises the following stages: model selection and pre-training, stratified data partitioning, evolutionary search, fitness evaluation, and fine-tuning with optimal configurations.

\subsection{Fine-tuning optimization problem}
\label{sec:3.1}
Given a pre-trained model $M = \{m_b: b \in \{0, \dots, B\}\} $, composed of $B+1$ sets of layers or blocks grouped based on their function (e.g., layers belonging to a residual block or a fully connected classifier), pre-trained on a source dataset with abundant labeled images $\mathcal{X}_s = \{(x_i^s, y_i^s)\}_{i=1}^{N_s} $, our goal is to determine an optimal fine-tuning configuration $\nu^* $ that maximizes the accuracy of the network on a target dataset $\mathcal{X}_t = \{(x_i^t, y_i^t)\}_{i=1}^{N_t} $, which contains different classes. The optimization problem can be framed as:

\begin{equation}
    \nu^* = \underset{\nu}{\arg\max} \, \text{Acc}(M(\omega^0), \lambda(\nu), \mathcal{X}_t)
\end{equation}

where $\omega^0$ represents the model pre-trained parameters, and $\lambda(\nu)$ defines the learning rates for each block during fine-tuning, given by:

\begin{equation}
\label{eq:2}
    \lambda(\nu) = \left\{ \eta_b(\nu) \lambda^0_b : b \in \{0, \dots, B\} \right\}
\end{equation}

where $\lambda^0_b $ is the predefined base learning rate for each block $b$, and $\eta_b(\nu) $ is a weight that adjusts the base learning rate according to the importance of each block during fine-tuning.
The function $\eta_b(\nu) $ controls the degree of fine-tuning for each set of layers and enables block-level selection (fine-tuning or freezing) and learning rate assignment for fine-tuning. For layers that are more crucial to adapting the model to the target dataset, $\eta_b(\nu) > 1 $ should be applied, effectively increasing the learning rate for those layers to facilitate more significant updates. Conversely, for less important layers, $\eta_b(\nu) < 1 $ should be used, reducing their learning rate to make only minimal adjustments. Layers that do not contribute to the adaptation process should have $\eta_b(\nu) = 0 $, freezing their parameters. This not only prevents them from being updated during training but also reduces computational costs and accelerates the fine-tuning process.

The optimal fine-tuning configuration $\nu^* $ is determined through an evolutionary search process that iteratively explores the configuration space.

\subsection{Evolutionary search}
\label{sec:3.2}

\begin{algorithm}[t]
    \SetAlgoLined
    \KwIn{Pre-trained model $M(w^0)$, Base learning rate $\lambda^0$, Population size $N_p$, Number of generations $N_g$, Number of Elites $N_e$, Fitness function $\Phi$}
    \KwOut{Optimal configuration $\nu^*$}
    Generate stratified data partitions \hfill (Eq. 16)\\
    \SetKwProg{Init}{Initialize}{:}{}
    \Init{Population $\mathcal{P}_0$}{
        Create $N_p$ random individuals $\nu^s$, $s \in \{0, \dots, N_p-1\}$ \dotfill (Eq. 1-3) \\
        Evaluate fitness $\Phi(\nu^s)$ for all individuals \dotfill (Eq. 4-8) \\
        Sort individuals by fitness \\
        $\nu^* \leftarrow$ individual with lowest fitness
    }
    \For{$g \gets 0$ \KwTo $N_g - 1$}{
        $\mathcal{M}_g \leftarrow \mathcal{P}_g$ \\
        \For{Elite individuals $\nu^s$, $s \in \{0, \dots, N_e-1\}$}{
            Exploit $\nu^s$ using random perturbations \dotfill (Eq. 9) \\
        }
        \For{$s \gets N_e$ \KwTo $N_p - 1$}{
            \If{$\mathcal{M}_g \neq \emptyset$}{
                Select parents $\nu^{P_a}, \nu^{P_b}$ and prototype $\nu^{P_*}$ from $\mathcal{M}_g$ \\
                $\nu^s \leftarrow$ Crossover($\nu^{P_a}, \nu^{P_b}$) \dotfill (Eq. 10) \\
                $\nu^s \leftarrow$ Mutate($\nu^s$) \dotfill (Eq. 13-14) \\
                $\nu^s \leftarrow$ Adopt($\nu^s, \nu^{P_a}, \nu^{P_b}, \nu^{P_*}$) \dotfill (Eq. 15) \\
                Clip genes of $\nu^s$ to [0,1] \\
                Evaluate fitness $\Phi(\nu^s)$ \dotfill (Eq. 4-8) \\
            } \Else {
                $\nu^s \leftarrow$ random individual \dotfill (Eq. 1-3) \\
                Evaluate fitness $\Phi(\nu^s)$ \dotfill (Eq. 4-8) \\
            }
        }
        $\mathcal{P}_{g+1} \leftarrow$ Elites $\cup$ Offspring \\
        Sort individuals in $\mathcal{P}_{g+1}$ by fitness \\
        Calculate extinction factors \dotfill (Eq. 12) \\
        $\nu^* \leftarrow$ individual with lowest fitness \\
        \If{Termination criteria fulfilled}{
            \textbf{break}
        }
    }
    Evaluate $\nu^*$ on target test set \\
    \Return{$\nu^*$}
    \caption{BioTune Algorithm}
    \label{alg:1}
\end{algorithm}

Fine-tuning configurations $\nu $ are encoded as genotypes, where the genotype of the $s $-th individual is represented as $\nu^s = \left[\nu^s_0, \nu^s_1, \dots, \nu^s_{B}, \nu^s_{B+1} \right]$. Each gene $\nu^s_b$, where $b \in {0, \dots, B}$, corresponds to an importance index for block $b$ and ranges from 0 to 1. The final gene, $\nu^s_{B+1} = \epsilon_f^s$, acts as a freezing threshold, differentiating blocks to be frozen from those to be fine-tuned. This threshold, $\epsilon_f^s$, is an optimization parameter that promotes diversity within the population \cite{dusan18novel}. By allowing this threshold to evolve, the algorithm can explore a wider range of sparsity in the fine-tuning configurations, preventing premature convergence to a suboptimal strategy where all layers are either tuned or frozen.

\begin{equation}
    \nu_b^{0,\dots,N_p-1} = U_{[0,1]} \quad \forall b = 0,\dots,B+1
\end{equation}

Each individual's fitness is evaluated using a function $\Phi$, which measures its predictive performance on a validation set. This evaluation comprises two stages: adaptive parameter setting and model performance sampling. Individuals are then ranked according to their fitness, and the top performers are selected as parents for the next generation. Evolutionary operations are then applied to these parents to generate offspring. The offspring are subsequently evaluated, and those with the highest fitness are selected to form the new population. This cycle repeats over multiple generations, progressively enhancing the population's overall fitness. The evolutionary process continues until the optimal fine-tuning configuration is identified, as indicated by the highest prediction accuracy (or lowest fitness value) on the validation set. This optimal configuration is then used to fine-tune the model with the full training set, which is subsequently evaluated on the held-out test set of the target dataset.

\subsubsection{Adaptive parameter setting}

For a given individual $\nu $ being evaluated, the learning rate weights $\eta(\nu_b) $ for each block $b $ are determined by the product of a selection mask $\mathcal{S} $ and an importance weight $\mathcal{W}$ vectors:

\begin{equation}
    \eta(\nu) = \mathcal{S} \cdot \mathcal{W}
\end{equation}

The selection mask $\mathcal{S} $ controls whether the corresponding set of layers is enabled for fine-tuning or remains frozen, based on the gene value $\nu_b $ and the freezing threshold $\epsilon_f$. Specifically, the mask is defined as:

\begin{equation}
    \mathcal{S}_b = \begin{cases} 
    0, & \text{if } \nu_b \leq \epsilon_f \\
    1, & \text{if } \nu_b > \epsilon_f 
    \end{cases}
\end{equation}

Here, $\mathcal{S}_b = 0 $ indicates that the layers in block $b $ are frozen, while $\mathcal{S}_b = 1 $ enables fine-tuning for that block. The threshold $\epsilon_f$ serves as the cutoff value to determine whether or not a block contributes to the fine-tuning process.
The importance weights $\mathcal{W} $ are calculated as follows:

\begin{equation}
    \mathcal{W}_b = 10^{2(\nu_b - 0.5)}
\end{equation}

This formulation allows $\mathcal{W} $ to exponentially scale the learning rates within [0.1,10], emphasizing layers with higher importance while de-emphasizing those with lower importance.
Finally, the learning rate $\lambda_b $ for each block $b $ is computed using Eq.~\ref{eq:2}, which integrates both the selection mask $\mathcal{S} $ and importance weight $\mathcal{W} $.

\subsubsection{Model performance sampling}

For each individual, the corresponding learning rate configuration is applied to the pre-trained model.  If $\eta(\nu_b)$ for a block is 0, its parameters are frozen. Otherwise, the block's learning rate is scaled by its corresponding weight $\eta(\nu_b)$, enabling differential parameter updates during training via gradient-based optimization.The model is then fine-tuned on the target training set for a fixed number of epochs using categorical cross-entropy loss:

\begin{equation}
    \mathcal{L} = -\frac{1}{N_t}\sum_{i=1}^{N_t} \sum_{c=1}^{C} y_{i,c} \log(\hat{y}_{i,c})
\end{equation}

where $N_t$ is the number of samples in the target dataset, $C$ is the number of classes, $y_{i,c}$ is the true label (1 if sample $i$ belongs to class $c$, 0 otherwise), and $\hat{y}_{i,c}$ is the predicted probability that sample $i$ belongs to class $c$.
To ensure robustness, the process is repeated with $N_s$ different random seeds. The average of the validation accuracies from these trials is used as the fitness metric, defined as:

\begin{equation} 
\label{eq:1} 
\Phi(\nu) = 1 - \frac{1}{N_s} \sum_{i=1}^{N_s} \psi_{\text{val}_i}(\nu)
\end{equation}

Here, $N_s $ represents the number of trials (seeds), and $\psi_{{val}_i} $ is the accuracy obtained on the validation set for the $i $-th trial. By using a negative sign, we convert this into a minimization problem, ensuring that the fitness $\Phi $ remains within the range $[0, 1] $. The population is then sorted according to their fitness values.

\subsubsection{Elitism and Exploitation}

To accelerate convergence and enhance performance, an exploitation operation is applied to the top $N_e $ \textit{elite} individuals, those with the highest fitness values in each generation. This process employs a stochastic perturbation approach, where a random modification is introduced to one of the genes in each \textit{elite} individual. After this adjustment, the individual's fitness is re-evaluated. If the modification results in improved fitness, the modified individual replaces the original. If no improvement is found, the original individual is preserved. The exploitation process is defined as:

\begin{equation}
    E_b = \begin{cases}
        \nu'_b = \nu_b + U_{[-1,1]} \cdot \delta \\
        g'_b = g_b + (\nu'_b-\nu_b)
    \end{cases}
\end{equation}

where $\delta $ is a fixed perturbation factor randomly scaled by a uniform distribution $U_{[-1,1]}$. This exploitation step allows the algorithm to further refine the \textit{elite} individuals without resorting to more computationally expensive optimization methods, thereby maintaining efficiency. A momentum term, $g$, is associated with each individual and updated to reflect changes in gene values.

\subsubsection{Crossover}
Non-elite individuals enter a mating pool for reproduction. Parent pairs are selected from this rank-based pool to produce offspring through gene combination. The crossover operation generates offspring genes through linear interpolation:

\begin{equation}
    \mathcal{C}_b = \begin{cases}
    \nu_b = I_{U_{[0,1]}} (\nu_b^{P_a},\nu_b^{P_b}) + g_b\\
    g_b = U_{[0,1]}g_b^{P_a}+U_{[0,1]}g_b^{P_b}
    \end{cases}
\end{equation}

\noindent where $I_{\alpha}(p,q)$ performs linear interpolation between values $p$ and $q$:

\begin{equation}
    I_{\alpha} (p,q) = \alpha p + (1-\alpha) q \quad \alpha \in [0,1]
\end{equation}

The $g_b$ term represents a momentum vector that is computed as a random linear combination of the parents' momentum vectors ($g_b^{P_a}$ and $g_b^{P_b}$). This momentum mechanism allows the offspring to inherit search directions from both parents, which helps guide the exploration of the fine-tuning configuration space while maintaining diversity. 

\subsubsection{Mutation}
\label{subsubsec:mutation}

Population diversity is maintained through an adaptive extinction operator $\zeta$, which measures each individual's relative fitness within the population \cite{starke19memetic}:

\begin{equation}
    \zeta^s = \frac{\Phi^s + \Phi_{\text{min}}(\frac{s}{N_p-1}-1)}{\Phi_{\text{max}}} \quad s \in \{0,\dots,N_p-1\}
\end{equation}

The mutation rate $\xi^s$ for each offspring is computed from its parents' extinction factors:

\begin{equation}
    \xi^s = \frac{1}{B+2}\left(\bar{\zeta}^P(B+1) + 1\right) \quad s \in \{0,\dots,N_p-1\}
\end{equation}

where $\bar{\zeta}^P = (\zeta^{P_1} + \zeta^{P_2})/2$ is the average parental extinction factor.

Each gene is mutated with probability $\xi^s$ according to:

\begin{equation}
    \mathcal{M}_b = \begin{cases}
    \nu'_b = \nu_b + \bar{\zeta}^P U_{[-1,1]} \\
    g'_b = g_b + (\nu'_b - \nu_b)
    \end{cases}
\end{equation}

The mutation magnitude is scaled by the average parental extinction factor $\bar{\zeta}^P$, and the momentum term $g_b$ is updated to reflect the gene modifications.

\subsubsection{Adoption}
The adoption process updates each individual's genes by incorporating information from both its parents and a high-ranking prototype individual ($P_*$). The update mechanism follows:

\begin{equation}
    \mathcal{A}_b = \begin{cases}
        \begin{aligned}
        \nu'_b &= \nu_b + I_{U_{[0,1]}}\\  
        &  \left(U_{[0,1]}\left(\frac{\nu_b^{P_a} + \nu_b^{P_b}}{2}-\nu_b\right), U_{[0,1]} \left(\nu_b^{P_*}-\nu_b\right)\right) 
        \end{aligned}\\
        g'_b = g_b + (\nu'_b-\nu_b)
    \end{cases}
\end{equation}

Based on Particle Swarm Optimization (PSO), this mechanism guides individuals toward promising regions of the search space by balancing parental inheritance with attraction toward successful prototypes. The momentum term $g_b$ is updated to reflect the gene modifications, maintaining search momentum.

\subsubsection{Termination criteria}

The evolutionary search terminates upon meeting either of two conditions. First, when the algorithm reaches the maximum allowed number of generations $N_g$. Second, when the best fitness value $\Phi_{\text{best}}$ shows no significant improvement over $N_c$ consecutive generations, formalized as $|\Phi_{\text{best}}^{(t)} - \Phi_{\text{best}}^{(t-N_c)}| < \epsilon_c$, where $t$ denotes the current generation and $\epsilon_c$ is a small convergence threshold value. Upon termination, the configuration achieving the lowest fitness value $\Phi_{\text{best}}$ is selected as the optimal solution.

\subsection{Stratified Data Partitioning}
\label{sec:3.3}

To address the computational demands of evolutionary search, we employ stratified data partitioning. The training dataset is divided into $N_s$ stratified folds, where:

\begin{equation}
    f_g = g \bmod N_s, \quad N_s \leq N_g
\end{equation}

Each fold preserves the original class distribution, and fold $f_g$ is used for evaluation at generation $g$. This rotating evaluation scheme serves dual purposes: reducing per-generation computational cost through smaller evaluation sets, and ensuring robust exploration by exposing candidates to different training samples across generations.The stratification ensures evaluations remain representative of the complete dataset characteristics while significantly reducing computational overhead. Although this rotating evaluation introduces variability in the fitness at each generation, it also acts as a form of regularization, preventing the optimization from overfitting to a single data-fold. The use of multiple seeds for each evaluation further ensures that the final selected configuration is robust and generalizes well to the entire dataset.

\subsection{Fine-tuning with optimal configuration}
\label{sec:3.4}
After the evolutionary search identifies the optimal configuration, the corresponding blocks are frozen, and the learning rates of the remaining trainable blocks are scaled.  The frozen blocks remain unchanged during fine-tuning, allowing the model to focus on adapting the active blocks' parameters.  This fine-tuning process, conducted on the entire training dataset, aims to maximize performance on the target task.  The final evaluation is performed on the held-out test set to assess the performance of the optimized model.

\section{Experimental setup}
\label{sec:4}

\subsection{Datasets}
\label{sec4:1}

To comprehensively evaluate the performance and generalization capabilities of our approach, we utilized a diverse set of image classification datasets. These datasets span a wide range of domains, including classification on digits, natural objects, fine-grained, and specialized domains. Table \ref{tab:1} presents an overview of the datasets used in our experiments, including the number of samples in the training and evaluation sets, as well as the number of classes for each dataset.

\begin{table}[H]
\centering
\renewcommand{\arraystretch}{1.2} 
\scalebox{1.0}{
\begin{tabular}{l c c c}
\cline{1-4}
Dataset        & Training  & Evaluation    & Classes \\ 
\hline
Flowers-102 \cite{nilsback08automated}     & 2040      & 6149          & 102 \\        
MNIST \cite{lecun98gradient}           & 60000     & 10000         & 10 \\        
USPS \cite{hull94database}           & 7291      & 2007          & 10 \\ 
SVHN \cite{netzer11reading}           & 73257     & 26032         & 10 \\
CIFAR-10\cite{krizhevsky09learning}        & 50000     & 10000         & 10 \\
STL-10  \cite{coates11analysis}        & 5000      & 8000          & 10 \\
FGVC-Aircraft \cite{maji13fine}   & 6667      & 3333          & 100 \\
DTD \cite{cimpoi14describing}            & 3760      & 1880          & 47 \\
ISIC2020 \cite{rotemberg21patient}        & 900       & 379           & 2 \\
\hline
\end{tabular}}
\caption{Distribution of samples for the experimental datasets}
\label{tab:1}
\end{table}

\subsection{Pre-trained model architecture}
\label{sec4:2}

\begin{figure}[t!]
  \centering
  \includegraphics[width=0.4\linewidth]{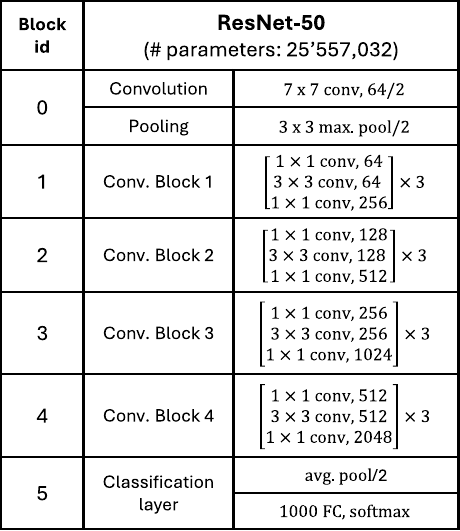}
  \caption{Overview of the ResNet-50 architecture, divided into six blocks used for fine-tuning and analysis.}
  \label{fig:3}
\end{figure}

To evaluate the classification performance of our proposed approach across datasets and to perform ablation studies, we utilize a ResNet-50 CNN architecture \cite{he16deep} pre-trained on the ImageNet dataset. ResNet-50 is a 50-layer deep convolutional neural network featuring shortcut connections that bypass certain layers, effectively mitigating the vanishing gradient problem and facilitating deeper network training. The architecture contains approximately 25.6 million parameters. For our experiments, we organize the ResNet-50 architecture into six functional blocks, as summarized in Fig.~\ref{fig:3}.

\subsection{Image preprocessing}
\label{sec4:3}

To prepare the input images, we first resize them to maintain their original aspect ratio, setting the shorter side to 232 pixels. Following resizing, we normalize the images using the mean and standard deviation values derived from the ImageNet dataset. Finally, each image is cropped to a standardized size of 224×224 pixels.
To focus specifically on assessing the impact of the fine-tuning process, we refrain from using any data augmentation techniques in this study. However, we acknowledge that employing higher-resolution images or data augmentation could potentially improve performance, as noted in prior research \cite{cui18large, cubuk19autoaugment}.

\subsection{BioTune implementation Details}
\label{sec4:4}

Our approach was implemented using the PyTorch 2.0 framework. All experiments were conducted on a high-performance workstation equipped with an AMD Ryzen Threadripper PRO 3995WX 2.7 GHz processor, 256 GB RAM, and an NVIDIA GeForce RTX A6000 GPU.

For the BioTune approach, we used a population size of \( N_p = 10 \), with \( N_e = 3 \) elite individuals, a maximum of \( N_g = 10 \) generations, and \( N_s = 3 \) random seeds per fitness evaluation to enhance robustness. The maximum number of epochs for BioTune training was set to 30, with a patience of 3 epochs for early stopping. A random perturbation of \( \delta = 0.25 \) was applied to elite individuals to encourage exploration. 

\subsection{Implementation of Benchmark Fine-tuning Methods}
\label{sec4:5}

To provide a comprehensive comparison, we implemented a range of well-established fine-tuning techniques, including both conventional and recent approaches \cite{davila24comparison}.

\begin{itemize}
    \item \textbf{Full Fine-tuning:} All layers of the pre-trained model are updated during training. This widely used method serves as a standard baseline for fine-tuning and allows for direct comparison with our approach.
    
    \item \textbf{Linear Probing \cite{kumar22fine}:} Only the final classification layer is trained, while all other layers remain frozen. This approach is computationally efficient but may limit the model’s ability to adapt fully to the new task.

    \item \textbf{$L^1$-SP Regularization \cite{li18explicit}:} Fine-tuning is guided by adding an \( L^1 \)-norm penalty on the difference between the pre-trained weights \( \omega_s \) and the fine-tuned weights \( \omega^0_s \) (\( || \omega_{s} - \omega^{0}_{s} ||_1 \)) to the objective function. This encourages the fine-tuned weights to remain close to the pre-trained values, promoting stable knowledge transfer.

    \item \textbf{$L^2$-SP Regularization \cite{li18explicit}:} Similar to \( L^1 \)-SP but using \( L^2 \)-norm regularization (\( || \omega_{s} - \omega^{0}_{s} ||_2^2 \)), which applies a smoother penalty on deviations from the pre-trained weights, reducing large parameter changes more gradually.

    \item \textbf{Gradual Unfreezing (Last$\rightarrow$First) \cite{howard18universal}:} Layers are progressively unfrozen during training, starting from the last layer and moving toward the first. This gradual adaptation allows the model to refine specific features before general ones, potentially improving task generalization.

    \item \textbf{Gradual Unfreezing (First$\rightarrow$Last) \cite{mukherjee20distilling}:} The reverse of the previous method, where layers are unfrozen progressively from the first to the last layer, aiming for a stepwise adaptation to new task features.

    \item \textbf{AutoRGN \cite{lee23surgical}:} Automatically adjusts learning rates for each layer at each training epoch based on each layer’s importance, calculated as the ratio of the gradient norm to the parameter norm. This dynamic adjustment prioritizes layers considered more relevant for the target task.

    \item \textbf{LoRA (Low-Rank Adaptation) \cite{hu21lora}:} This method introduces low-rank matrices within the original weights of a pre-trained model. Instead of updating the full weight matrix, LoRA fine-tunes only the low-rank matrices for each layer, significantly reducing the number of trainable parameters. Although originally developed for transformer models, we adapted LoRA for convolutional neural networks (CNNs) by incorporating low-rank matrices into each CNN layer for targeted fine-tuning.
\end{itemize}

\section{Experimental Results}
\label{sec:5}

\subsection{Performance on diverse image classification datasets}
\label{sec5:1}

We conduct a comprehensive evaluation of the proposed BioTune model across a diverse set of image classification tasks summarized in Table \ref{tab:1}. The experimental results are presented in Table~\ref{tab:2}, where we utilize test set accuracy as the primary performance metric. We report the mean accuracy across three independent runs along with standard errors (shown in parentheses) to account for training variations and assess result stability. For BioTune specifically, we evaluate the top-5 performing configurations identified across all explored generations and report the highest achieved accuracy.

\begin{table*}[tb]
    \centering
    \renewcommand{\arraystretch}{1.4} 
    \caption{Comparison of accuracy with standard errors across three runs for various fine-tuning methods, evaluated on different datasets. The top score for each dataset is highlighted in bold. For BioTune, the second line shows the relative performance improvement compared to FT.}
    \scalebox{0.82}{
    \begin{tabular}{r | c c c | c c | c c | c c}
        \cline{1-10}
                    & \multicolumn{3}{c|}{Digits}  	& \multicolumn{2}{c|}{Objects} & \multicolumn{2}{c|}{Fine-grained} & \multicolumn{2}{c} {Specialized}    \\ 
        Method & MNIST         & USPS          &SVHN           & CIFAR-10  & STL-10       & Flowers-102    & FGVC-Aircraft  & DTD       & ISIC2020   \\ \hline
        FT          & 98.96 (0.0)  & 97.05 (0.1)  & 95.56 (0.2)  & 95.65 (0.1)  & 97.33 (0.0)  & 85.33 (0.5)  & 58.68 (1.9)  & 68.03 (0.1)  & 78.91 (0.7)  \\
        \cline{1-10}
        LP          & 92.53 (0.2)  & 91.78 (0.0)  & 44.08 (0.1)  & 81.03 (0.0)  & 96.86 (0.1)  & 82.72 (0.6)  & 33.61 (0.4)  & 66.00 (0.1)  & 77.49 (0.5)  \\
        $L^1$-$SP$  & 98.98 (0.1)  & 97.31 (0.0)  & 96.01 (0.0)  & 95.60 (0.0)  & 97.01 (0.0)  & 87.82 (0.5)  & 60.55 (1.9)  & 68.52 (0.1)  & 80.62 (1.2)  \\
        $L^2$-$SP$  & 98.87 (0.0)  & 97.00 (0.1)  & 95.47 (0.1)  & 95.78 (0.1)  & 97.20 (0.1)  & 85.29 (0.5)  & 61.56 (1.1)  & 69.01 (0.2)  & 79.77 (2.0)  \\
        G-LF        & 98.82 (0.1)  & 96.72 (0.2)  & 94.00 (0.0)  & 93.77 (0.0)  & 97.32 (0.0)  & 87.59 (0.9)  & 54.77 (1.3)  & 67.85 (0.3)  & 77.49 (1.2)  \\
        G-FL        & 98.57 (0.0)  & 96.86 (0.1)  & 94.91 (0.0)  & 95.43 (0.0)  & 97.01 (0.0)  & 86.14 (0.2)  & 49.22 (0.7)  & 65.42 (0.3)  & 76.92 (1.3)  \\
        AutoRGN     & 99.00 (0.0)  & 96.91 (0.2)  & \B{96.08} (0.0)  & 96.05 (0.0)  & 96.92 (0.1)  & 85.5 (0.3)  & 57.94 (0.8)  & 65.70 (0.2)  & 79.48 (0.4)  \\
        LoRA        & 98.51 (0.1)  & 96.92 (0.0)  & 95.46 (0.1)  & 95.17 (0.1)  & 97.46 (0.1)  & 86.01 (0.2)  & 54.78 (1.3)  & 68.17 (0.4)  & 80.91 (1.0)  \\
        \cline{1-10}
        BioTune & \B{99.13} (0.0) & \B{97.57} (0.1) & 95.85 (0.0) & \B{96.09} (0.1) & \B{97.50} (0.0) & \B{91.68} (0.1) & \B{64.40} (0.6) & \B{69.27} (0.6) & \B{82.90} (0.8) \\[-1.5ex]
        (Ours) & \footnotesize{+0.2\%} & \footnotesize{+0.5\%} & \footnotesize{+0.3\%} & \footnotesize{+0.5\%} & \footnotesize{+0.2\%} & \footnotesize{+6.7\%} & \footnotesize{+9.7\%} & \footnotesize{+1.8\%} & \footnotesize{+5.1\%} \\[1pt]
        \hline
    \end{tabular}}
    \label{tab:2}
\end{table*}

The experimental results demonstrate that BioTune consistently achieves competitive or superior performance compared to existing fine-tuning methods across most datasets, with SVHN being the only exception where AutoRGN slightly outperforms our approach. For datasets closely aligned with the source domain of natural images (Digits and Objects categories), BioTune shows moderate but consistent improvements over the FT baseline, ranging around 0.5\%. While these gains may appear modest, they are significant given the already high baseline performance.

The most substantial improvements are observed on fine-grained and specialized datasets. In fine-grained classifications (Flowers-102 and FGVC-Aircraft), we observe significant gains, with FGVC-Aircraft achieving a 9.7\% improvement over FT. Specialized datasets, which include target domains quite different from the source domain, DTD (textures) and ISIC2020 (dermoscopy), demonstrate high performance gains of 1.8\% and 5.1\% respectively, highlighting BioTune's effectiveness in handling significant domain shifts.

Among the benchmark fine-tuning methods, performance varies considerably across datasets, demonstrating that existing strategies may be effective for certain distribution shifts but potentially detrimental for others. For instance, AutoRGN excels on several datasets (MNIST, SVHN, CIFAR-10, Flowers-102, ISIC2020) but shows significant performance degradation on others (DTD, STL-10). Regularization-based strategies ($L^1$-SP and $L^2$-SP) and LoRA demonstrate stronger performance on fine-grained specialized datasets but struggle to improve accuracy on datasets closer to the source domain.

\begin{figure}[t!]
  \centering
  \includegraphics[width=0.7\linewidth]{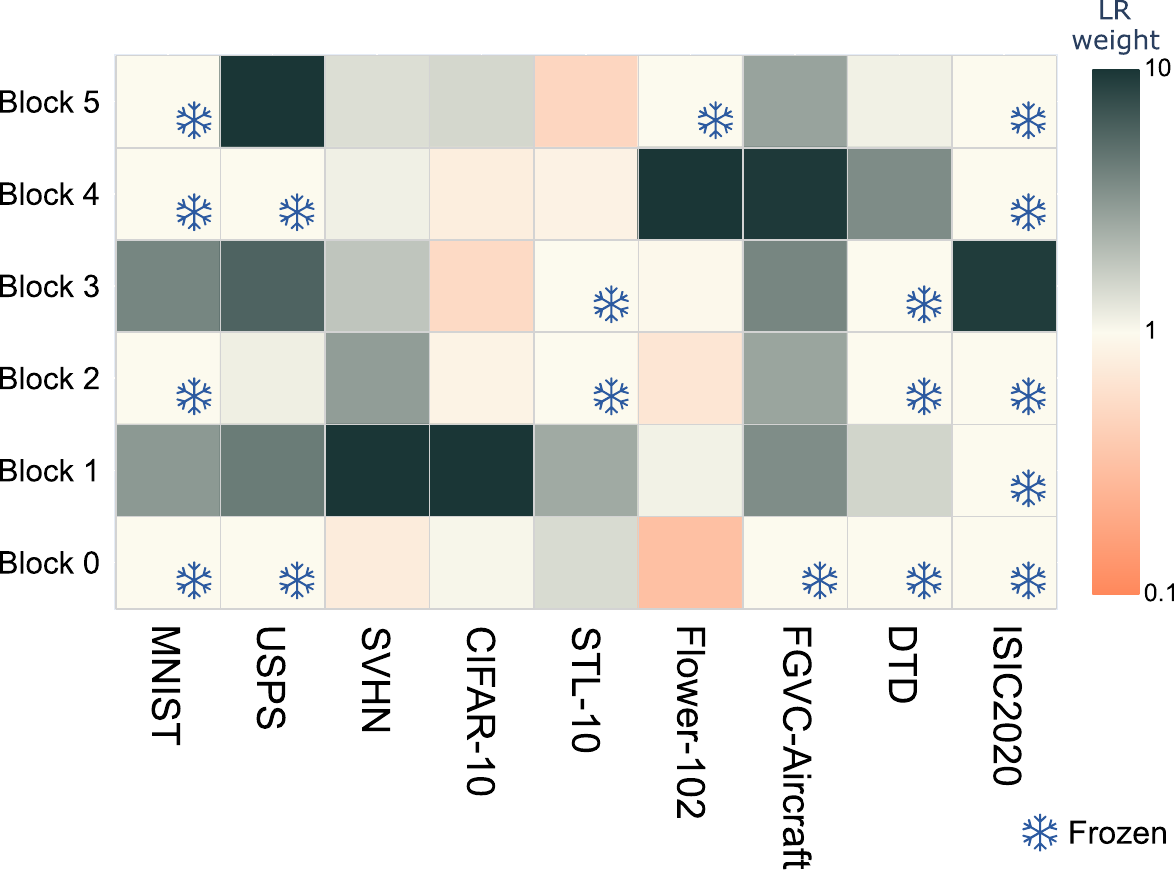}
  \caption{Optimal fine-tuning configurations discovered by BioTune across datasets. Frozen layers are marked with a snowflake symbol, and learning rate weights are shown as a heatmap (0.1 to 10).}
  \label{fig:4}
\end{figure}

Figure~\ref{fig:4} provides a comprehensive visualization of the optimal fine-tuning configurations discovered by BioTune for each dataset. In this visualization, frozen layers are denoted by a snowflake symbol, while the learning rate weights are represented as a heatmap with values ranging from 0.1 to 10. The analysis reveals distinct patterns in layer-wise adaptation across different types of transfer learning scenarios.

For datasets that closely align with the source domain (e.g., MNIST, CIFAR-10), BioTune tends to assign higher weights to the bottom layers of the network, which are known to extract general, low-level features such as edges and textures. Conversely, for fine-grained classification tasks (Flowers-102, FGVC-Aircraft) and specialized domains (DTD, ISIC2020), the model prioritizes adaptation of the top layers through higher learning rates. These empirical findings align with \cite{lee23surgical}, which suggests that input-level distribution shifts benefit more from adjusting lower layers closer to the input, while output-level shifts (such as fine-grained classifications) are better addressed by modifying the top layers that capture more task-specific features.

Table~\ref{tab:3} presents the percentage of trainable parameters utilized by BioTune relative to the total available parameters in ResNet-50 for each dataset. Our analysis reveals significant variations in parameter utilization across different tasks, demonstrating BioTune's ability to automatically determine the optimal subset of parameters for fine-tuning.

\begin{table*}[tb]
    \centering
    \renewcommand{\arraystretch}{1.4} 
    \caption{Percentage of trainable parameters with respect to the original total number of available trainable paramters in ResNet-50 for each dataset}
    \scalebox{1.0}{
    \begin{tabular}{r | c c c | c c | c c c c}
        \cline{1-10}
                    & \multicolumn{3}{c|}{Digits}  	& \multicolumn{2}{c|}{Objects} & \multicolumn{4}{c}{Specialized}    \\ 
        Method & MNIST         & USPS          &SVHN           & CIFAR-10  & STL-10       & Flower-102    & FGVC-Aircraft  & DTD       & ISIC2020   \\ \hline
        BioTune  & 29.97  & 36.86  & 100.0   & 100.0   & 64.93   & 99.12   & 99.96   & 64.89   & 29.93   \\
        \hline
    \end{tabular}}
    \label{tab:3}
\end{table*}

Some tasks, such as SVHN, CIFAR-10, Flowers-102, and FGVC-Aircraft, require extensive model adaptation with nearly all parameters being fine-tuned ($>$99\%). In contrast, other tasks achieve optimal performance with substantially reduced parameter sets, MNIST and ISIC2020, which utilize less than 30\% of the available parameters. Intermediate cases like STL-10 and DTD show moderate parameter utilization around 65\%. By selectively freezing certain layers and avoiding unnecessary gradient computations and backpropagation, BioTune achieves significant computational efficiency while maintaining or improving performance. This characteristic could be particularly valuable in resource-constrained scenarios or when rapid adaptation to new tasks is required.

\subsection{Performance accross various network architectures}
\label{sec5:2}

\begin{figure}[t!]
  \centering
  \includegraphics[width=0.95\textwidth]{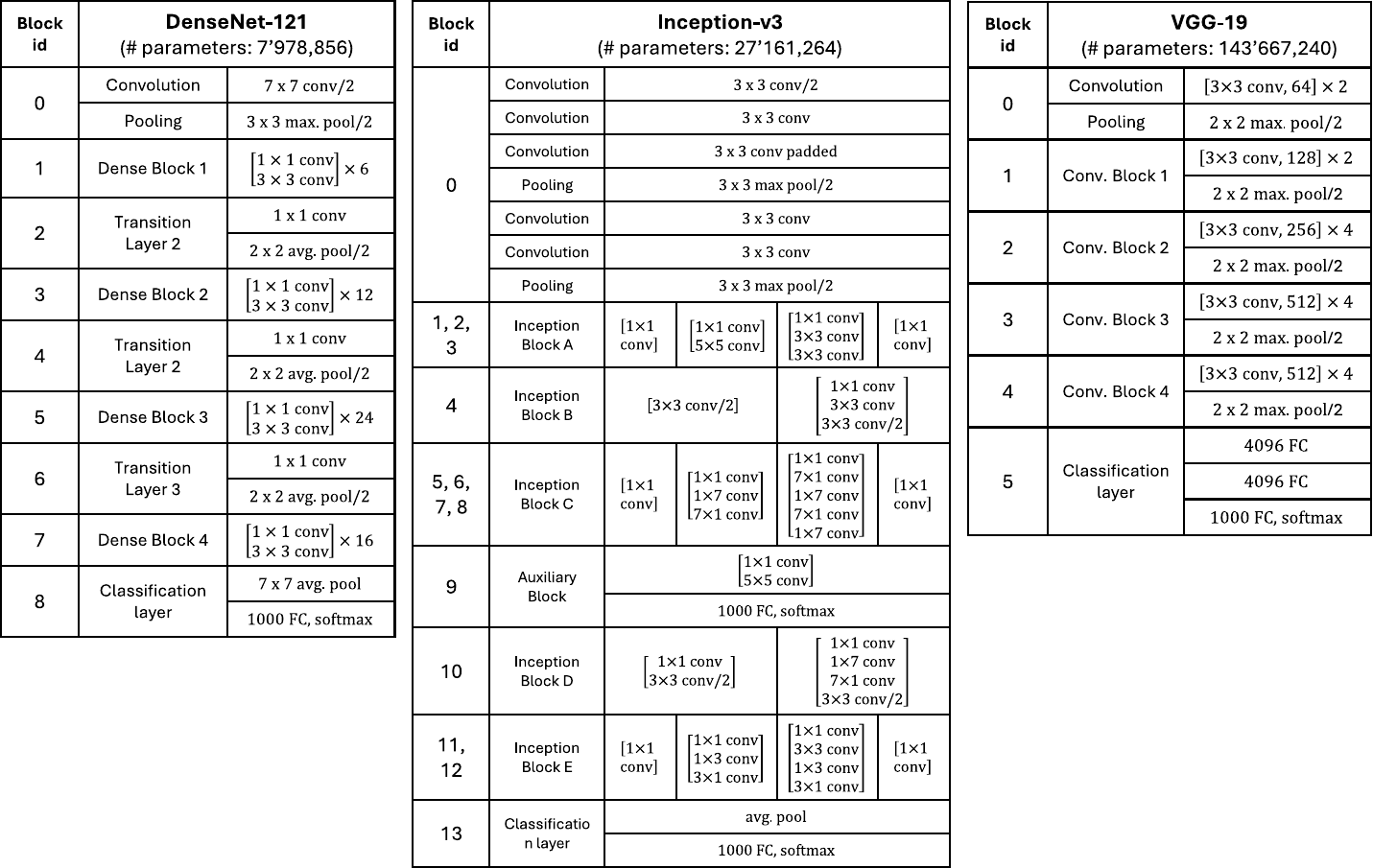}
  \caption{Network architectures and blocks.}
  \label{fig:5}
\end{figure}

To evaluate BioTune versatility and effectiveness across different model architectures, we extend our analysis beyond ResNet-50 to include three additional widely-used architectures: DenseNet-121 \cite{huang17densely}, VGG16 \cite{simonyan15deep}, and Inception-v3 \cite{szegedy15going}. These architectures were selected to represent diverse designs, varying in model size, connectivity patterns, and computational requirements. For each architecture, we group layers into functional blocks as represented in Figure~\ref{fig:5}.

We conduct our comparative analysis on the Flowers-102 dataset, employing all fine-tuning methods described in Section \ref{sec4:5}. Following Section~\ref{sec5:1}, we report mean accuracies from three independent runs along with standard errors. The comprehensive results are presented in Table \ref{tab:4}.

\begin{table}[tbp]
    \centering
    \renewcommand{\arraystretch}{1.4} 
    \caption{Comparison of accuracy with standard errors for the Flowers-102 dataset across three runs for various architectures. The top score for each architecture is highlighted in bold. For BioTune, the second line shows the relative performance improvement compared to FT.}
    \scalebox{1.0}{
    \begin{tabular}{r | c c c c }
        \cline{1-5}
                    & \multicolumn{4}{c}{Network}  	\\ 
        Method   & ResNet-50    & Densenet-121  & VGG-19       & Inception-v3   \\ \hline
        FT          & 85.33 (0.5)   & 90.62 (0.1)   & 66.22 (1.4)  & 85.42 (0.4)  \\
        \cline{1-5}
        LP          & 82.72 (0.6)   & 87.73 (0.6)   & 66.10 (0.9)  & 71.47 (0.4)    \\
        $L^1$-$SP$  & 87.82 (0.5)   & 88.63 (0.1)   & 53.26 (2.6)  & 84.61 (0.4)    \\
        $L^2$-$SP$  & 85.29 (0.5)   & 90.67 (0.0)   & 65.82 (0.5)  & 85.01 (0.1)    \\
        G-LF        & 87.59 (0.9)   & 90.21 (0.0)   & 66.39 (0.4)  & 88.65 (0.3)    \\
        G-FL        & 86.14 (0.2)   & 82.41 (0.2)   & 64.06 (0.6)  & 85.23 (0.6)    \\
        AutoRGN     & 85.5 (0.3)    & 88.64 (0.1)   & 63.25 (1.0)  & 81.58 (1.1)    \\
        LoRA        & 86.01 (0.2)   & 90.23 (0.3)   & 66.05 (2.3)  & 82.95 (0.5)    \\
        \cline{1-5}
        BioTune & \B{91.08} (0.1) & \B{92.04} (0.1) & \B{69.7} (1.7)   & \B{89.38} (0.1) \\[-1.5ex]
        (Ours)  & \footnotesize{+6.7\%}        & \footnotesize{+3.4\%}       & \footnotesize{+5.3\%}       & \footnotesize{+4.64\%}        \\[1pt]
        \hline
    \end{tabular}}
    \label{tab:4}
\end{table}

The experimental results presented in Table~\ref{tab:4} demonstrate BioTune's consistently superior performance across all evaluated architectures. Our method achieves significant improvements over standard fine-tuning (FT), with performance gains ranging from 3\% to 6\%. The most substantial improvement is observed with ResNet-50 (+6.7\%), while DenseNet-121, despite showing the smallest relative improvement (+3.4\%), achieves the highest absolute accuracy of 92.04\%. Among baseline methods, only Gradual Layer Freezing (G-LF) shows improvements across most cases, though not matching BioTune's performance.

\begin{figure}[t!]
  \centering
  \includegraphics[width=0.6\linewidth]{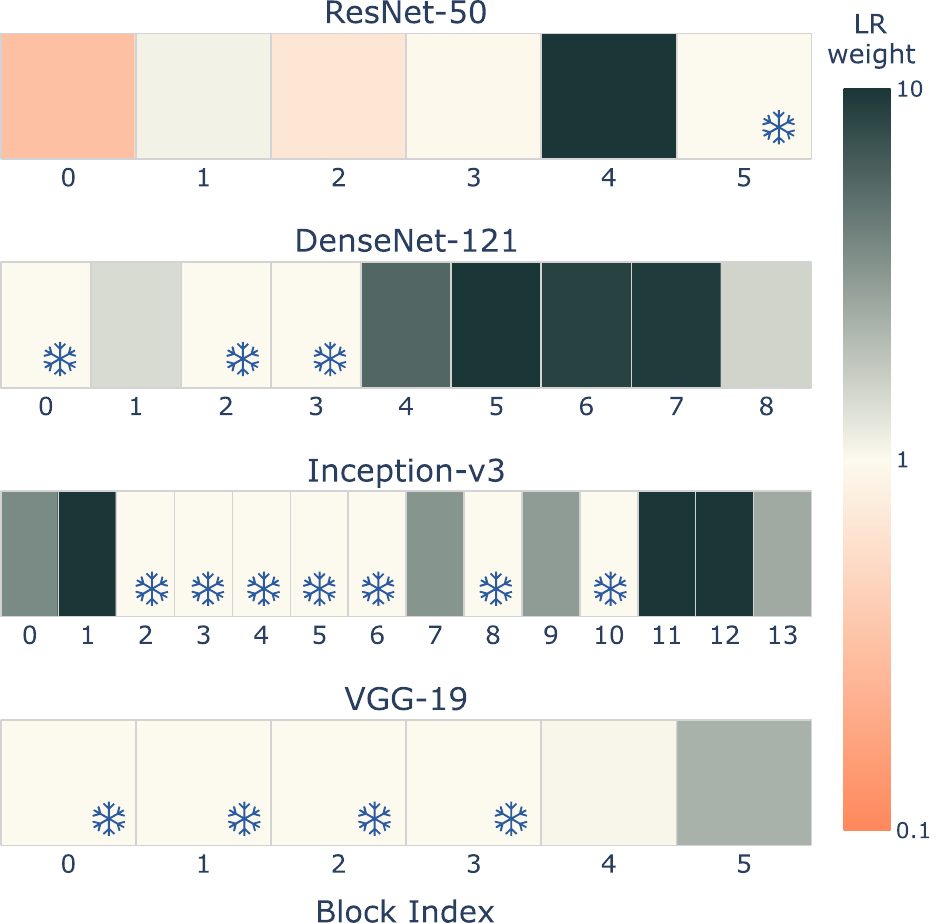}
  \caption{Optimal fine-tuning configurations for each architecture, as identified by BioTune.}
  \label{fig:6}
\end{figure}

Figure~\ref{fig:6} visualizes the optimal fine-tuning configurations discovered by BioTune for each architecture. A consistent pattern emerges across architectures: higher learning rates are assigned to the top layers, aligning with our findings from Section~\ref{sec5:1} regarding the importance of top-layer adaptation for fine-grained classification tasks. Notably, most architectures exhibit frozen intermediate blocks, with the exception of ResNet-50 which maintains more active layers throughout. Inception-v3 displays a unique pattern with elevated weights in both the initial and final blocks, suggesting that low-level general features extracted by the initial CNN layers may strongly influence the subsequent Inception modules. 

Table~\ref{tab:5} presents the percentage of trainable parameters utilized by the optimal BioTune configurations across architectures. The analysis reveals distinct patterns of parameter utilization: ResNet-50 requires near-complete model adaptation (99.12\%), followed by VGG-19 (92.44\%). In contrast, DenseNet-121 and Inception-v3 achieve optimal performance with substantially reduced parameter sets, utilizing 86.36\% and 66.26\% respectively. The notably lower parameter utilization in Inception-v3 suggests its architecture may enable more efficient transfer learning through selective parameter updating.

\begin{table}[tbp] 
    \centering
    \renewcommand{\arraystretch}{1.4} 
    \caption{Percentage of trainable parameters utilized by BioTune relative to total available parameters across network architectures}
    \scalebox{1.0}{
    \begin{tabular}{r | c c c c }
        \cline{1-5}
        Method   & ResNet-50    & Densenet-121   & Inception-v3  & VGG-19   \\ \hline
        BioTune       & 99.12    &  86.36     &  66.26 & 92.44 \\
        \hline
    \end{tabular}}
    \label{tab:5}
\end{table}

\subsection{Ablation studies}

This section presents a series of ablation studies designed to investigate the impact of various components and hyperparameters of BioTune on its overall performance. These studies provide insights into the effectiveness of individual design choices and contribute to a deeper understanding of the method's behavior.

\subsubsection{Optimization effectiveness compared with other bio-inpired approaches}

\begin{figure}[t!]
  \centering
  \includegraphics[width=0.85\linewidth]{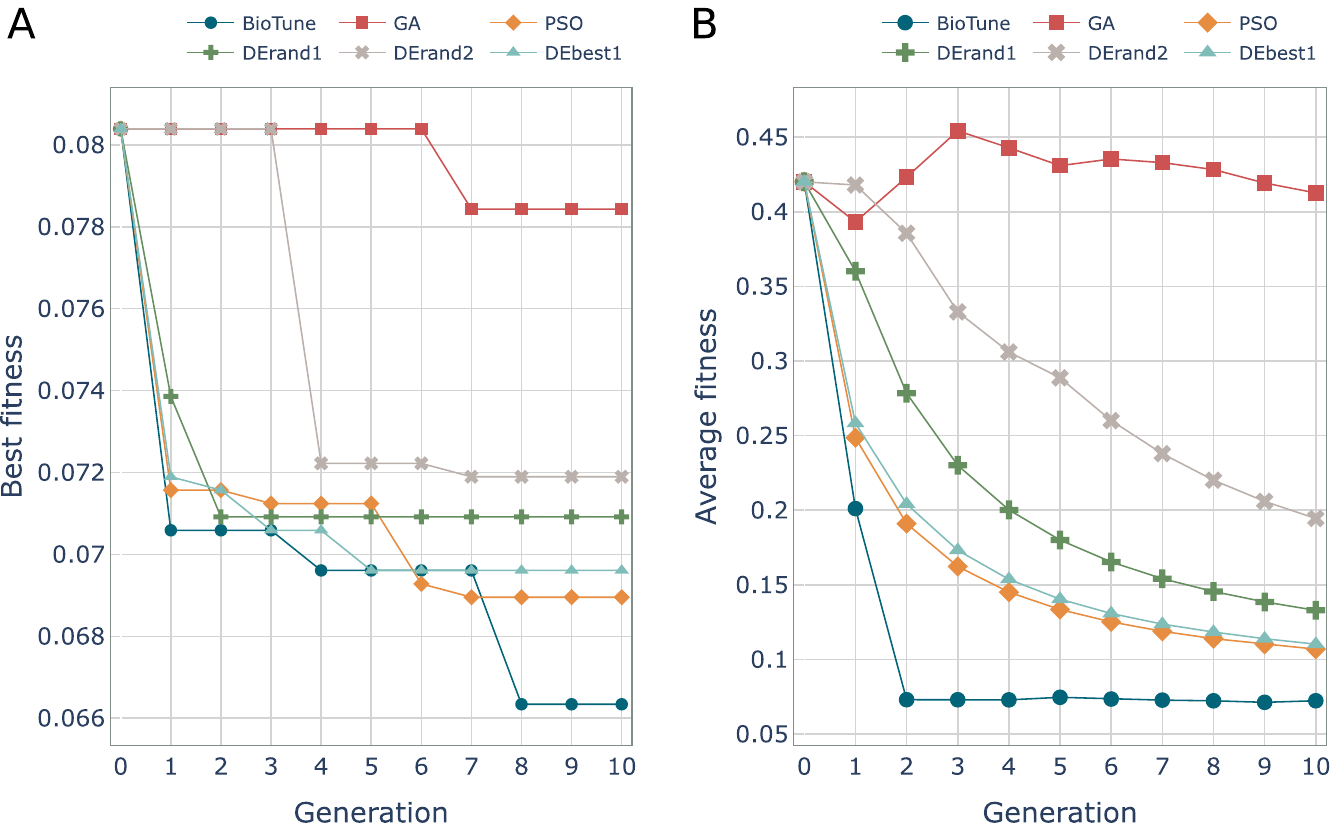}
  \caption{Convergence analysis of bio-inspired optimization approaches. \textbf{A.} Best fitness evolution across generations. \textbf{B.} Average population fitness dynamics.}
  \label{fig:7}
\end{figure}

We evaluate BioTune's optimization performance against several established bio-inspired algorithms: Genetic Algorithms (GA), multiple variants of Differential Evolution (DE/rand/1, DE/best/1, DE/rand/2, DE/best/2), and Particle Swarm Optimization (PSO). For a fair comparison, all algorithms were initialized with identical random populations and implemented using the NiaPy framework \cite{vrbancic18niapy}. Each algorithm tackles the same fine-tuning optimization problem with hyperparameters empirically tuned for optimal performance. Figure \ref{fig:7} illustrates the convergence dynamics through best and average fitness trajectories across generations.

The results demonstrate that most evolutionary methods achieve rapid initial fitness improvements, with convergence typically occurring within 8 generations, followed by fitness plateaus. BioTune consistently outperforms all evaluated bio-inspired optimizers, achieving superior fitness scores. PSO and DE/best/1 appear as the next most effective approaches, while GA exhibits comparatively lower performance. Notably, BioTune's average fitness trajectory reveals rapid population convergence toward optimal individuals, leveraging collective population momentum to explore the solution space more efficiently than alternative methods.

\begin{figure}[t!]
  \centering
  \includegraphics[width=0.8\linewidth]{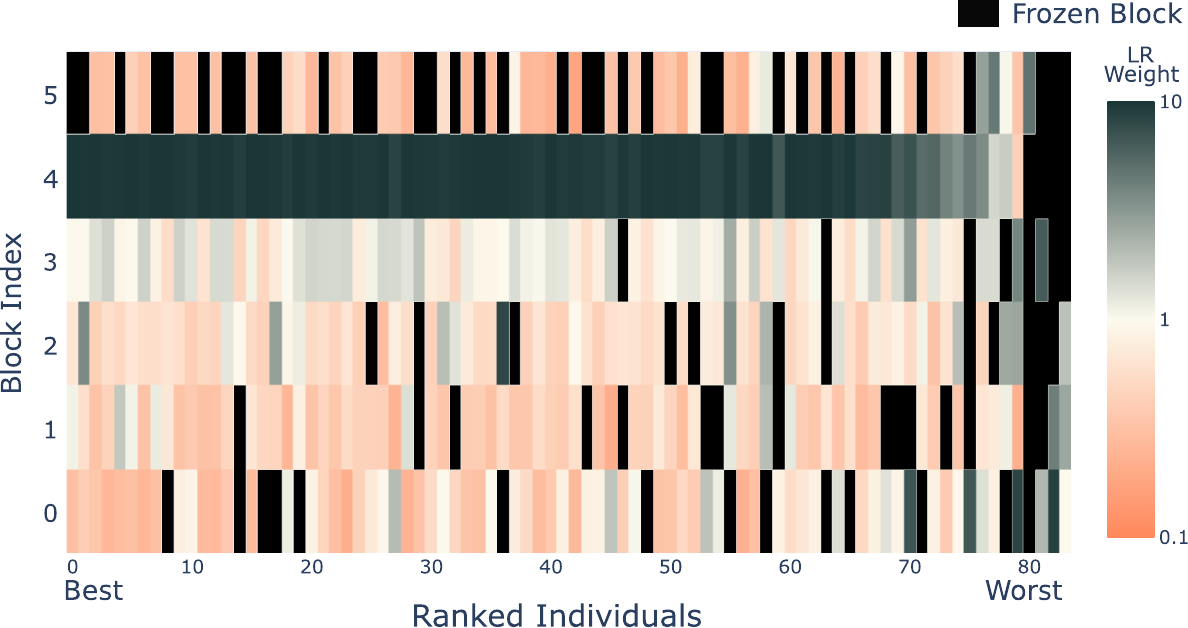}
  \caption{Visualization of all individuals generated by BioTune, ranked by performance. Black regions represent frozen blocks, while the heatmap (0.1--10) shows weight values.}
  \label{fig:8}
\end{figure}

Figure~\ref{fig:8} visualizes all individuals generated during the optimization process. Frozen blocks are shown in black, while weights are displayed as a heatmap ranging from 0.1 to 10. Individuals are sorted from best on the left to worst on the right. The distribution reveals distinct patterns in BioTune's exploration strategy.
The poorest-performing individuals exhibit high variability in both block freezing patterns and weight distributions, indicating the algorithm's initial broad exploration phase. As performance improves, configurations adapt to more selective freezing patterns, with top performers converging on minimal freezing, focused on final layers. This visualization demonstrates BioTune progressively refining configurations from diverse initial explorations to optimal fine-tuning strategies.

\subsubsection{Population and elite size}
\begin{figure}[t!]
  \centering
  \includegraphics[width=0.55\linewidth]{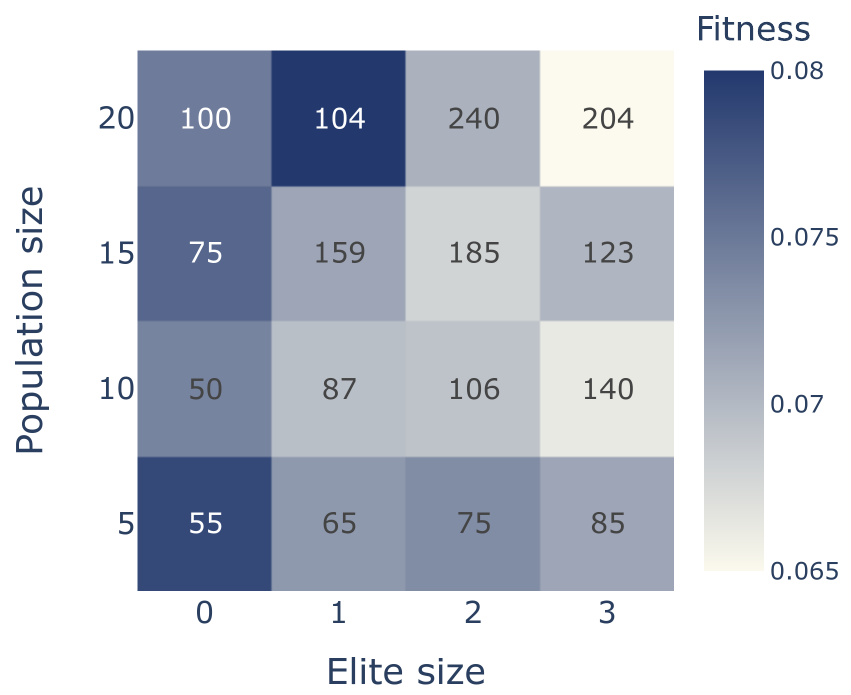}
  \caption{Performance analysis across different population and elite size configurations. Cell colors indicate best fitness achieved, while numbers inside cells show the total fitness computations required.}
  \label{fig:9}
\end{figure}

We analyzed the relationship between BioTune's performance and its population and elite size parameters. Performance was evaluated using two metrics: computational efficiency (measured by fitness computation count) and solution quality (best fitness achieved). Our experiments covered population sizes from 5 to 20 individuals and elite sizes ranging from 0 (no elitism) to 3 elite individuals. Figure \ref{fig:9} presents a heatmap visualization of these results.

The top-performing configurations are predominantly located on the right side, indicating that increasing population and elite sizes generally enhances fitness performance. However, this improvement comes at the cost of a significant increase in the number of fitness computations. It is worth noting that not all generations are necessary to achieve optimal results, as the algorithm will terminate if no further improvement is found after a certain number of generations. This adaptive termination criterion can lead to lower numbers of fitness evaluations, even for larger population and elite sizes. Balancing performance and computational cost, we selected a population size of 10 and an elite size of 3 for our experiments.

\subsubsection{Importance weight function}

We evaluated four different importance weight functions $\mathcal{W}(\nu)$ to determine their impact on BioTune's performance:

\paragraph{Discriminative} This binary approach only implements selective freezing of blocks, maintaining the base learning rate $\lambda(\nu) = \mathcal{S}(\nu) \cdot \lambda^0$ for fine-tuning:
\begin{equation}
     \mathcal{W}_b = 1 \quad \text{for } b = 0, \dots, B
\end{equation}

\paragraph{Scaled} The importance weight scales each gene value relative to the maximum gene value:
\begin{equation}
    \mathcal{W}_b = \frac{\nu_b}{\max(\boldsymbol{\nu})} \quad \text{for } b = 0, \dots, B
\end{equation}

where $\max(\boldsymbol{\nu})$ represents the maximum element of the genome, excluding the freezing threshold $\epsilon_f$.

\paragraph{Normalized} The importance weight is normalized to [0, 1] using the freezing threshold $\epsilon_f$:
\begin{equation}
    \mathcal{W}_b = \frac{\nu_b - \epsilon_f}{\max(\boldsymbol{\nu}) - \epsilon_f} \quad \text{for } b = 0, \dots, B
\end{equation}

\paragraph{Exponential} The importance weight for each block is calculated using an exponential function, mapping gene values to the range [0.1, 10] and allowing for broader exploration of learning rates:
\begin{equation}
    \mathcal{W}_b = 10^{2(\nu_b - 0.5)} \quad \text{for } b = 0, \dots, B
\end{equation}

\begin{table}[tb]
\centering
\renewcommand{\arraystretch}{1.2} 
\caption{Fitness values for different importance weight functions. Lower values indicate better performance.}
\scalebox{1.0}{
\begin{tabular}{l  c }
\cline{1-2}
Evol. Method            & Fitness           \\ \hline
Discriminative          & 0.122 $\pm$ 0.0   \\ 
Scaled                  & 0.121 $\pm$ 0.001 \\ 
Normalized              & 0.119 $\pm$ 0.003 \\ 
Exponential  & \bf{0.069} $\pm$ 0.002 \\ 
\hline
\end{tabular}}
\label{tab:6}
\end{table}

Table~\ref{tab:6} summarizes the fitness performance for each weight function. The Normalized Exponential weight function significantly outperforms other approaches, achieving a fitness of 0.069 compared to approximately 0.12 for the alternatives. This suggests that exponentially scaling the importance weights enables BioTune to explore a broader range of learning rates, ultimately finding more optimal configurations. 

\subsubsection{Fitness function}
We evaluate BioTune's performance using three different fitness functions:

\paragraph{Acc} The average validation accuracy is used as the performance metric:

\begin{equation}
    \Phi_{\text{Acc}} = 1 - \frac{1}{N_s} \sum_{i=1}^{N_s} \psi_{\text{val}_i}
\end{equation}

where $\psi_{val}$ represents the validation accuracy and $N_s$ the number of seeds evaluated for each individual.

\paragraph{AccStd} The average validation accuracy is adjusted by the standard deviation among the seeds. This fitness function aims to reduce variability and dependence on individual seeds:

\begin{equation}
    \Phi_{\text{AccStd}} = 1 - \frac{1}{N_s} \sum_{i=1}^{N_s} \psi_{\text{val}_i} + \sigma(\psi_{\text{val}})
\end{equation}

\paragraph{Loss} The validation loss is used instead of validation accuracy as the performance measurement:

\begin{equation}
    \Phi_{\text{Loss}} = 1 - \mathcal{L}_{val}
\end{equation}

\begin{figure}[t!]
  \centering
  \includegraphics[width=0.6\linewidth]{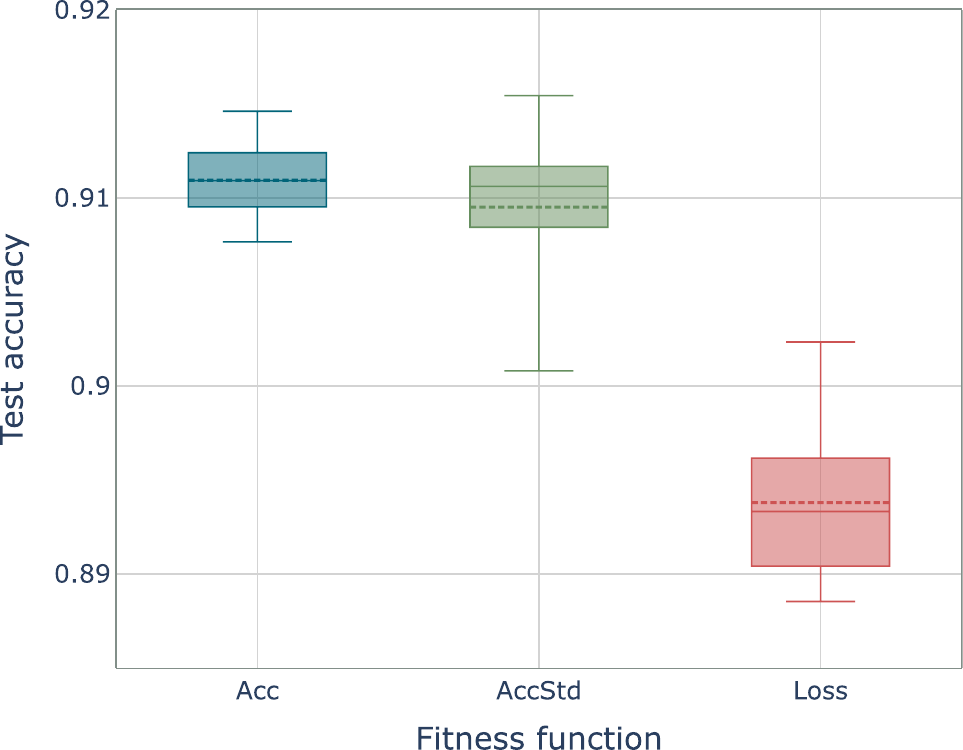}
  \caption{Test accuracy across three fitness functions. Acc: Validation accuracy. AccStd: Validation accuracy corrected with standard deviation. Loss: Validation loss.}
  \label{fig:10}
\end{figure}

Figure \ref{fig:10} presents test accuracy results for the top-5 individuals found with each fitness function. The Acc fitness function achieves the highest median test accuracy (91.1\%) with the smallest interquartile range, indicating good and consi  stent performance. Loss exhibits the worst performance, with test accuracy below 90\%. AccStd shows similar performance to Acc but with higher variability. Consequently, we use Acc for our experiments.

\subsubsection{Percentage of training data per generation}

\begin{figure}[t!]
  \centering
  \includegraphics[width=0.6\linewidth]{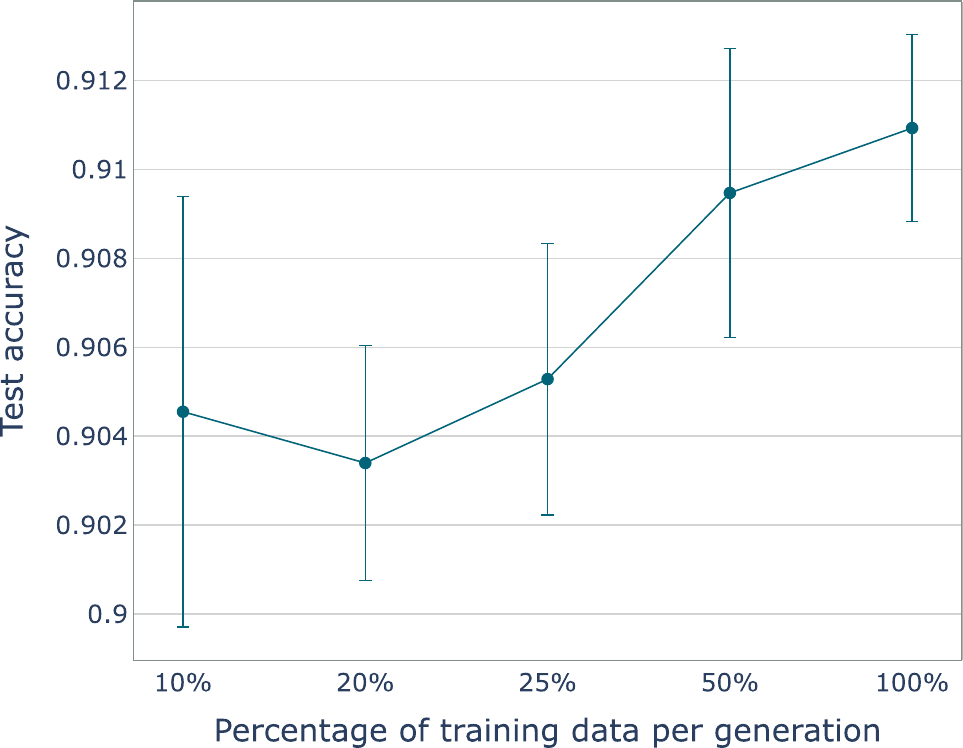}
  \caption{Test accuracy of top-5 individuals across different training data percentages. Error bars indicate standard deviation.}
  \label{fig:11}
\end{figure}

We explored how varying the percentage of training data per generation affects BioTune's performance. This analysis highlights the trade-off between computational cost and performance gains when using smaller data subsets. Figure \ref{fig:11} shows the test accuracy for the top-5 individuals across different training data percentages, with error bars indicating the standard deviation. Table \ref{tab:7} summarizes the mean test accuracy and computation time for optimizing the Flowers-102 dataset.

Our results demonstrate that high test accuracy ($>$90\%) is achievable using only 10\% of the training data, with a minimal computation time (1.6 hours). Although increasing training data per generation improves performance and reduces variance, it also increases computational costs. The peak accuracy of 91.1\% using 100\% of the training data requires 11.4 hours of computation. For our experimental validation, we selected a balanced 50\% split to optimize the accuracy-efficiency trade-off. Although comparing FLOPs with baselines is complex, Table~\ref{tab:7} indicates that evolutionary search requires a one-time computational effort. This results in a reusable, optimized fine-tuning setup, with costs adjustable based on the percentage of data used in the search.

\begin{table}[tb]
\centering
\renewcommand{\arraystretch}{1.2} 
\caption{Mean test accuracy and computation time for different training data percentages}
\scalebox{1.0}{
\begin{tabular}{l  c c  c c c}
\cline{1-6}
                    & \multicolumn{5}{c}{Percentage of Training dataset}  \\ 
Dataset        
& 10\%          & 20\%          & 25\%          & 50\%          & 100\%    \\ \hline
Mean Accuracy       & 90.46   & 90.34     & 90.53    & 90.95    & 91.1 \\ 
Comp. time (hours)    & 1.6   & 2.2     & 3.7    & 6.0    & 11.4 \\ 
\hline
\end{tabular}}
\label{tab:7}
\end{table}

\section{Conclusions}
\label{sec:6}

In this paper, we introduced BioTune, a novel evolutionary-based adaptive fine-tuning approach designed to optimize transfer learning in image classification tasks. BioTune dynamically determines layer freezing and learning rates, effectively adapting pre-trained models to new target domains, even with limited labeled data. Our comprehensive benchmarking across diverse datasets and architectures demonstrated BioTune's superior performance compared to existing fine-tuning methods, consistently achieving higher accuracy and improved efficiency by selectively freezing layers and reducing trainable parameters.
However, our work has some limitations. Although BioTune demonstrated strong performance on a range of tasks, its evaluation on other model families, such as Vision Transformers (ViTs), remains an area of future work. Adapting block-based optimization to the architectural specifics of ViTs would strengthen its generalizability. Additionally, integrating BioTune with advanced data augmentation techniques and exploring its potential synergy with other transfer learning approaches, such as domain adaptation, could further enhance performance.
Future research could explore extending BioTune to other deep learning tasks beyond image classification, such as object detection, semantic segmentation, and natural language processing. Investigating the interpretability of the learned fine-tuning configurations could provide valuable insights into the transfer learning process and guide the design of more efficient architectures. We believe that BioTune can be a useful tool for the development of more robust and adaptable deep learning models, opening the path for more effective transfer learning across a wider range of applications and domains. 

\bibliographystyle{ieeetr} 
\bibliography{biblio.bib}

\section*{Acknowledgments}
This work was supported in part by the Japan Science and Technology Agency (JST) CREST including AIP Challenge Program under Grant JPMJCR20D5, and in part by the Japan Society for the Promotion of Science (JSPS) Grants-in-Aid for Scientific Research (KAKENHI) under Grant 25K21247.

\end{document}